# A Comparative Study of Recent Large Language Models on Generating Hospital Discharge Summaries for Lung Cancer Patients

Yiming Li, MS, [1] Fang Li, PhD, [2] Kirk Roberts, PhD, [1] Licong Cui, PhD, [1] Cui Tao, PhD, [2] Hua Xu, PhD, [3]*

*corresponding author

[1]McWilliams School of Biomedical Informatics, The University of Texas Health Science Center at Houston, Houston, TX 77030, USA

[2]Department of Artificial Intelligence and Informatics, Mayo Clinic, Jacksonville, FL 32224, USA

[3]Department of Biomedical Informatics and Data Science, School of Medicine, Yale University, New Haven, CT 06510, USA

**Corresponding Author:**

Hua Xu, PhD, Department of Biomedical Informatics and Data Science, School of Medicine, Yale University, New Haven, CT 06510, USA; email: hua.xu@yale.edu

# ABSTRACT

**Objective**

Generating discharge summaries is a crucial yet time-consuming task in clinical practice, essential for conveying pertinent patient information and facilitating continuity of care. Recent advancements in large language models (LLMs) have significantly enhanced their capability in understanding and summarizing complex medical texts. This research aims to explore how LLMs can alleviate the burden of manual summarization, streamline workflow efficiencies, and support informed decision-making in healthcare settings.

**Materials and Methods**

Clinical notes from a cohort of 1,099 lung cancer patients were utilized, with a subset of 50 patients for testing purposes, and 102 patients used for model fine-tuning. This study evaluates the performance of multiple LLMs, including GPT-3.5, GPT-4, GPT-4o, and LLaMA 3 8b, in generating discharge summaries. Evaluation metrics included token-level analysis (BLEU, ROUGE-1, ROUGE-2, ROUGE-L) and semantic similarity scores between model-generated summaries and physician-written gold standards. LLaMA 3 8b was further tested on clinical notes of varying lengths to examine the stability of its performance.

**Results**

The study found notable variations in summarization capabilities among LLMs. GPT-4o and fine-tuned LLaMA 3 demonstrated superior token-level evaluation metrics, while LLaMA 3 consistently produced concise summaries across different input lengths. Semantic similarity scores indicated GPT-4o and LLaMA 3 as leading models in capturing clinical relevance.

**Conclusion**

This study contributes insights into the efficacy of LLMs for generating discharge summaries, highlighting LLaMA 3's robust performance in maintaining clarity and relevance across varying clinical contexts. These findings underscore the potential of automated summarization tools to enhance documentation precision and efficiency, ultimately improving patient care and operational capability in healthcare settings.

# INTRODUCTION

Writing discharge summaries is a time-consuming process for healthcare providers, with a mean of 8.1 minutes to dictate a summary and a median of 29.2 minutes to transcribe it, resulting in an average of 3.6 pages per summary [1], [2]. Recent advances in natural language processing (NLP), particularly through the development of large language models (LLMs), have demonstrated significant potential in enhancing and automating clinical documentation [3], [4]. LLMs such as GPT-4 and LLaMA 3 are at the forefront of these advancements [5], [6]. GPT-4, developed by OpenAI, features 1.76 trillion parameters and supports processing up to 25,000 words at once, allowing it to handle contexts eight times larger than those managed by GPT-3 [7]. Its advanced variant, GPT-4o, extends these capabilities, offering enhanced performance in text understanding [8]. Similarly, LLaMA3, with its 8 billion and 70 billion parameter versions, represents a substantial leap in the LLaMA series [9]. Leveraging extensive training on diverse datasets, including medical literature and clinical notes, these models hold the promise of generating coherent and contextually accurate discharge summaries, potentially transforming the documentation process in healthcare settings [3].

The application of LLMs to automate the generation of clinical documents, such as discharge summaries, has been explored in a few studies. For example, Singh's study assessed ChatGPT's ability to create ophthalmic discharge summaries and operative notes using 24 prompts based on various subspecialties [10]. The model produced responses quickly, tailored to input quality, with discharge summaries being valid but containing generic text and operative notes requiring significant tuning [10]. ChatGPT admitted and corrected mistakes promptly, indicating potential for positive healthcare impact with focused training and human verification [10]. Clough et al. evaluated the feasibility of using AI to produce high-quality hospital discharge summaries by comparing summaries generated by ChatGPT with those written by junior doctors [11]. They used 25 mock patient vignettes to create 50 discharge summaries, 25 from each source, and assessed them for quality and adherence to a minimum dataset [11]. The AI-generated summaries were deemed acceptable by GPs 100% of the time, compared to 92% for junior doctor summaries, and GPs struggled to distinguish between AI and human-written summaries [11].Dubinski investigated the use of ChatGPT for writing neurosurgical discharge summaries and operative reports, comparing it with the currently employed speech recognition software at a major university hospital [12]. The study found that using ChatGPT significantly reduced the time required to compose these documents for three neurosurgical diseases, though the factual correctness for craniotomy reports was lower [12]. The results indicate that while ChatGPT can efficiently assist in medical writing, further exploration of its optimal use and ethical considerations is necessary [12]. Ruinelli et al. conducted a comparative study on the automatic generation of hospital discharge summaries in an Italian-speaking region, using ChatGPT [13]. They evaluated the potential of ChatGPT to draft discharge summaries that could be reviewed and finalized by medical experts [13]. Despite some shortcomings, the study found that ChatGPT-produced summaries were relatively close in quality to those generated by humans, even in Italian [13].

In this study, we focus on lung cancer patients as a use case for applying LLMs to discharge summary generation. Lung cancer, the leading cause of cancer-related deaths worldwide, presents a particularly complex clinical scenario due to the extensive medical interventions and frequent transitions between care facilities [18], [19],[20],[21],[22]. The implications of this study are significant for both clinical and informatics domain. It presents the potential to enhance patient care by ensuring that detailed and accurate discharge summaries are generated more efficiently, thus allowing medical practitioners to focus more on direct patient care. From the informatics perspective, this study demonstrates the practical application of LLMs in real-world healthcare settings, potentially paving the way for broader adoption of these artificial intelligence tools to address similar challenges in clinical documentation.

## METHODS

The framework (Figure 1) for generating discharge summaries integrates advanced technologies to streamline documentation for lung cancer patients with COVID-19. Patient records, including diagnostic reports, lab tests, imaging studies, and clinical notes, are preprocessed using OCR technology to convert scanned documents into textual information. NLP with LLMs, further analyzing on specific medical datasets, is then employed to extract key information and generate comprehensive, contextually accurate summaries. This process ensures that structured and unstructured data are effectively combined, resulting in detailed discharge summaries.

## Dataset

The dataset for this study was sourced from Memorial Hermann Hospital in Texas, one of the largest and most comprehensive health systems in the region, renowned for its high standards of patient care and advanced medical practices. During the period of January 2021 to May 2021, we collected a cohort of all patients with COVID-19 tested. From the collection, we further identified 1,099 lung cancer patients. This dataset includes a diverse array of medical report types, such as lab test results, radiology reports, progress notes, and admission notes. The records cover a wide range of clinical scenarios and diagnostic outcomes, providing a rich and varied dataset for analysis.

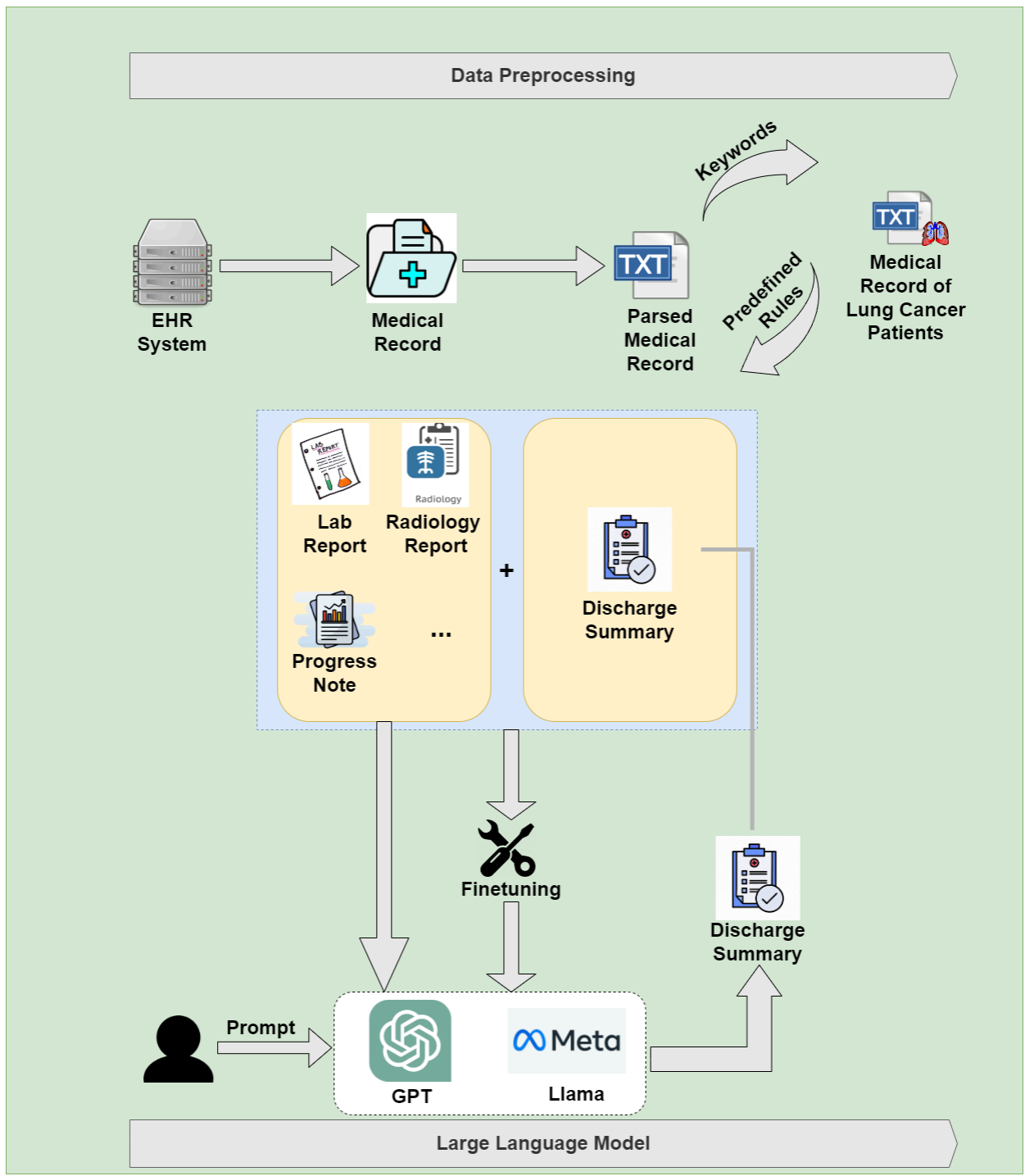

Figure 1　Overview of the framework

## Data preprocessing

In this study, we focused on preprocessing data for generating discharge summaries for lung cancer patients using advanced NLP techniques. We included patients diagnosed with lung cancer and used keyword searches to identify relevant patient records. The query looked for terms related to various types of neoplasms combined with terms related to lung anatomy.

The keywords used in the search were:

- "neoplasm" or "tumor" or "cancer" or "malignan-" or "carcinoma" or "sarcoma"
- "lung" or "bronchus" or "lobe"

The clinical notes for the selected patients were exported as PDFs, and OCR was used to convert these PDFs into text.

## Models

### GPT

GPT series represents a groundbreaking advancement in NLP [35], [36], [37], [38]. The latest iteration, GPT-4, boasts an impressive 1.76 trillion parameters and can handle up to 32,000 tokens at once, compared to GPT-3's limit of 4,096 tokens [40], [41, p. 4], [42]. This extended token capacity allows it to perform a wide range of tasks, from language translation and summarization to question answering[43], [44], [45]. A further optimized version, GPT-4o, features a more sophisticated architecture and fine-tuning capabilities, enhancing its performance in specialized tasks[46]. These advancements make GPT-4 and GPT-4o highly valuable tools for applications like text generation, chatbots, and research in fields such as medicine and finance [39].

### LLaMA

LLaMA (Large Language Model Meta AI), is a state-of-the-art language model designed for cutting-edge natural language understanding and generation [47], [48]. LLaMA 2 introduced a robust architecture with improved training algorithms, making it effective for various NLP tasks [47]. Expanding on this progress, the latest iteration, LLaMA 3 launched 8 billion and 70 billion parameter versions [49]. These advancements position LLaMA 3 as a leading NLP model, with impactful applications across healthcare, education, and business.

## Experiment Setup

### Dataset Split

We focused on clinical notes within 8,000 tokens (n=152) to meet the context length requirements for comparing the performance of different LLMs in discharge summary generation, using notes from 50 patients for testing and notes from 102 patients for further fine-tuning the LLaMA 3 model. Additionally, we utilized clinical notes from additional 947 lung cancer patients to assess the effect of varying context length on LLaMA 3's performance.

## Model Parameters

In this study, we utilized several large language models, including GPT-3.5-turbo-16k, GPT-4, GPT-4o, and LLaMA 3. The parameters used for these models are as follows:

For the GPT models, we used the default parameters provided by OpenAI. The temperature was set to 1.0, controlling the randomness of the model's outputs, with lower values resulting in more deterministic responses. The maximum tokens were set to 16,384 for GPT-3.5-turbo-16k and up to 8192 for GPT-4, determining the maximum number of tokens in the generated output; the specific limit can vary based on the model variant. The top-p (nucleus sampling) parameter was set to 1.0, controlling the diversity of the output by considering the cumulative probability of token sequences, with a value of 1.0 indicating no nucleus sampling. The frequency penalty was set to 0.0 to reduce verbosity by penalizing repeated tokens in the output, and the presence penalty was also set to 0.0 to enhance diversity by penalizing repeated themes or topics. No stop sequences were specified, allowing the model to generate text until it reached the maximum token limit or another stopping criterion.

For the LLaMA 3 model, we employed custom parameters to optimize its performance. We used the LLaMA 3 8b variant, which indicates the specific model size and variant used. The maximum sequence length (max_seq_len) was set to 8196 tokens, determining the maximum length of the input sequence that the model can process. Additionally, the maximum batch size (max_batch_size) was set to 6, which specifies the maximum number of sequences that can be processed in a single batch.

These parameters were chosen to optimize the performance of each model for generating high-quality discharge summaries for lung cancer patients.

## Prompt

For our experiments, we used the following prompts to generate discharge summaries:

> *"{"role": "system", "content": "You are an assistant good at writing discharge summary for lung cancer patients"},*
> *{"role": "user", "content": "Could you generate a discharge summary for this lung cancer patient:[clinical notes]".*

These prompts were designed to instruct the language models to focus specifically on generating high-quality discharge summaries tailored for lung cancer patients based on the provided clinical notes.

The pretrained GPT models were conducted experiments on a server featuring 8 Nvidia A100 GPUs, each providing 80GB of memory. In contrast, the pretrained LLaMA models underwent fine-tuning and inference on a server equipped with 5 Nvidia V100 GPUs, each offering a memory capacity of 32GB.

## Fine-tuning

In this study, we conducted fine-tuning on the LLaMA 3 8b model to optimize its performance for generating discharge summaries tailored for lung cancer patients. The fine-tuning process involved adjusting several key parameters. We set r to 16 and specified target modules including "q_proj", "k_proj", "v_proj", "o_proj", "gate_proj", "up_proj", and "down_proj". Additionally, lora_alpha was set to 16 with lora_dropout at 0, and bias was optimized as "none".

To manage memory efficiently and accommodate longer context dependencies, we employed use_gradient_checkpointing set to "unsloth". The fine-tuning process was initialized with random_state set to 3407, ensuring reproducibility across experiments. We did not utilize rank stabilized LoRA (use_rslora) or LoftQ (loftq_config), focusing instead on optimizing the core parameters critical for enhancing model performance in discharge summary generation for complex medical scenarios like lung cancer.

## The Effect of Context Length

In this study, we conducted an analysis on the effect of context length using the LLaMA 3 8b model for generating discharge summaries tailored for lung cancer patients. To evaluate this, we divided the patient cohort into five groups based on the length of their clinical notes in tokens: less than 50,000 tokens, 50,000-100,000 tokens, 100,000-200,000 tokens, 200,000-500,000 tokens, and greater than 500,000 tokens.

Limited by the maximum context length of 8,196 tokens, we devised an iterative prompt strategy to ensure comprehensive coverage of all clinical notes for each patient. The prompts used in our study were structured as follows:

Initial Prompt Set:

> *[*
> *{"role": "system", "content": "You are an assistant good at writing discharge summary for lung cancer patients"},*
> *{"role": "user", "content": "Could you generate a discharge summary for this lung cancer patient? I have all the clinical notes for these patients, but they are extensive. I'll provide them to you in multiple prompts."},*
> *],*

Iterative Prompts:

    *{"role": "user", "content": f"Here is a portion of the previous clinical notes, which you should use as the basis to generate the discharge summary. Please do not generate the discharge summary until I confirm that all clinical notes have been provided: {clinical notes}"},*
    *],*

Final Prompt Set:
    *[*
    *{"role": "system", "content": "You are an assistant good at writing discharge summary for lung cancer patients"},*
    *{"role": "user", "content": "All clinical notes for this patient have been provided. Please generate the discharge summary for this lung cancer patient based on the provided information"},*
    *],*

This prompt methodology enabled us to systematically manage and feed the extensive clinical notes to the LLaMA 3 8b model, ensuring that it could effectively generate accurate and comprehensive discharge summaries tailored to each patient's medical history and condition.

## Evaluation

In this study, we employed a comprehensive evaluation framework to assess the quality and accuracy of discharge summaries generated by various models. Our evaluation metrics included Bilingual Evaluation Understudy (BLEU), Recall-Oriented Understudy for Gisting Evaluation (ROUGE)-1, ROUGE-2, and ROUGE-L, which are widely used for token-level evaluation in natural language processing tasks. BLEU measures the precision of generated text by comparing n-grams (typically up to 4-grams) with reference texts, providing insights into how well the generated summaries match the reference standards. ROUGE-1 evaluates overlap in unigrams (individual words), while ROUGE-2 extends this to bigrams (pairs of consecutive words), and ROUGE-L considers the longest common subsequence (LCS) between the generated and reference texts, rewarding longer matching sequences. Additionally, we measured semantic similarity to capture the overall meaning and coherence of the generated discharge summaries compared to the gold standards [50].

## **RESULTS**

Figure 2 provides a detailed comparison of the token counts for the gold standard summaries, and the outputs generated by various large language models. The gold standard summaries, which serve as the benchmark, have a significantly lower average token count of 1562.52 with a standard deviation of 783.405, reflecting a more concise and focused summarization.

The outputs from the large language models show notable variations. GPT-3.5 and GPT-4 generate relatively shorter summaries, with token counts averaging 705.9 and 754.12, respectively. In contrast, the fine-tuned version of GPT-4 (GPT-4o) produces more detailed summaries with an average token count of 1043.92. LLaMA 3 generates the most concise summaries, averaging 624.26 tokens, while its fine-tuned version produces summaries that are closest to the gold standard in terms of length, with an average token count of 1625.79. These results highlight differences in the summarization capabilities of each model and their efficiency in condensing clinical information.

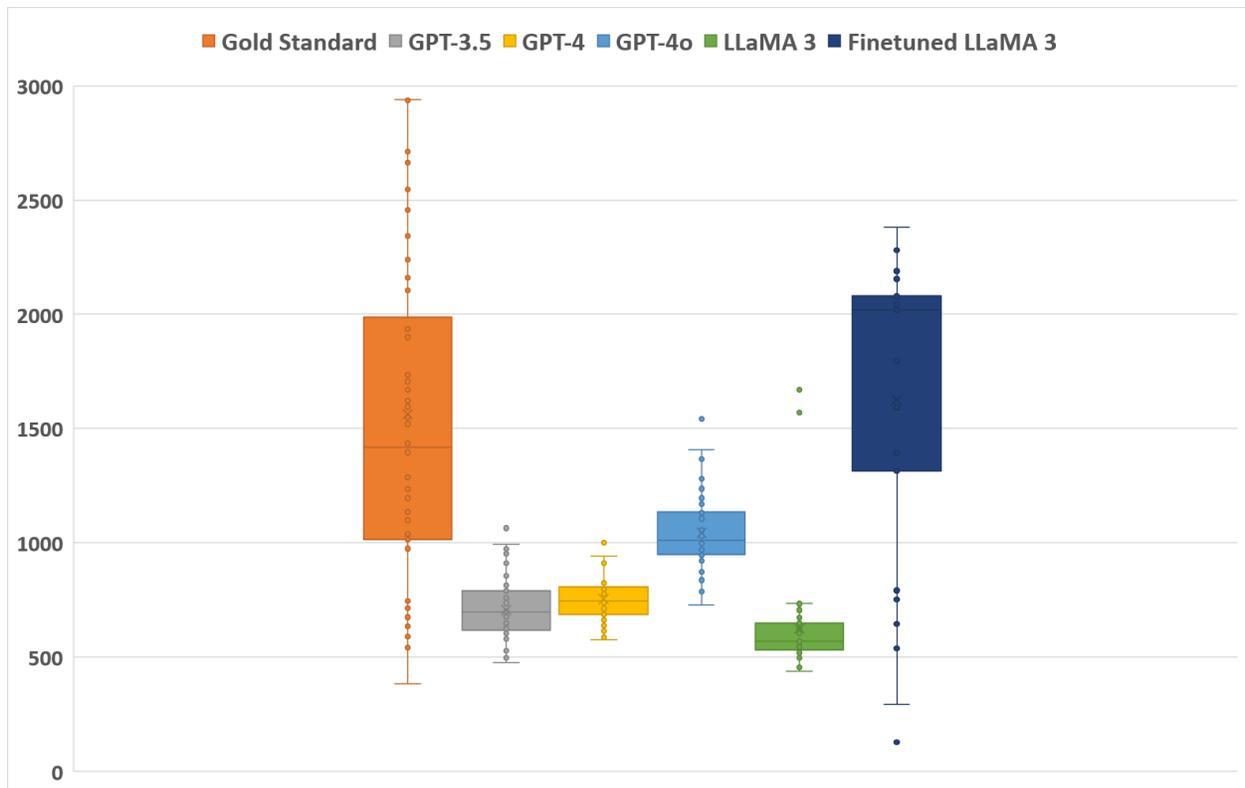

Figure 2 Box plot for the number of tokens in gold standard and outputs of large language models.

Figure 3 presents a comparison of the performance metrics for the generated summaries from GPT-3.5, GPT-4, GPT-4o, LLaMA 3, and the fine-tuned LLaMA 3 models. The metrics include BLEU, ROUGE-1, ROUGE-2, ROUGE-L, and semantic similarity scores, which are crucial for evaluating the quality and coherence of the summaries.

The BLEU scores are generally low across all models, with the fine-tuned LLaMA 3 achieving the highest score at 0.04. ROUGE-1 scores are highest for GPT-4o at 0.38, indicating better overlap of unigrams with the gold standard summaries. ROUGE-2 scores, which consider bigram overlap, show GPT-4o leading with a score of 0.11, closely followed by GPT-4 and the fine-tuned LLaMA 3 at 0.1. For ROUGE-L, GPT-4o and the fine-tuned LLaMA 3 both score 0.16, suggesting these models generate more structurally similar summaries to the gold standard.

In terms of semantic similarity, which assesses the semantic coherence of the summaries, the LLaMA 3 achieves the highest score at 0.837, followed by GPT-4o at 0.83. GPT-4 and GPT-3.5 also perform well with scores of 0.82 and 0.81, respectively. Fine-tuned LLaMA 3, however, has a significantly lower semantic similarity score of 0.63, indicating a weaker alignment with the gold standard summaries.

These results highlight that while all models perform variably across different metrics, the GPT-4o consistently show better performance in terms of ROUGE and semantic similarity scores, suggesting their superior capability in generating high-quality, coherent summaries.

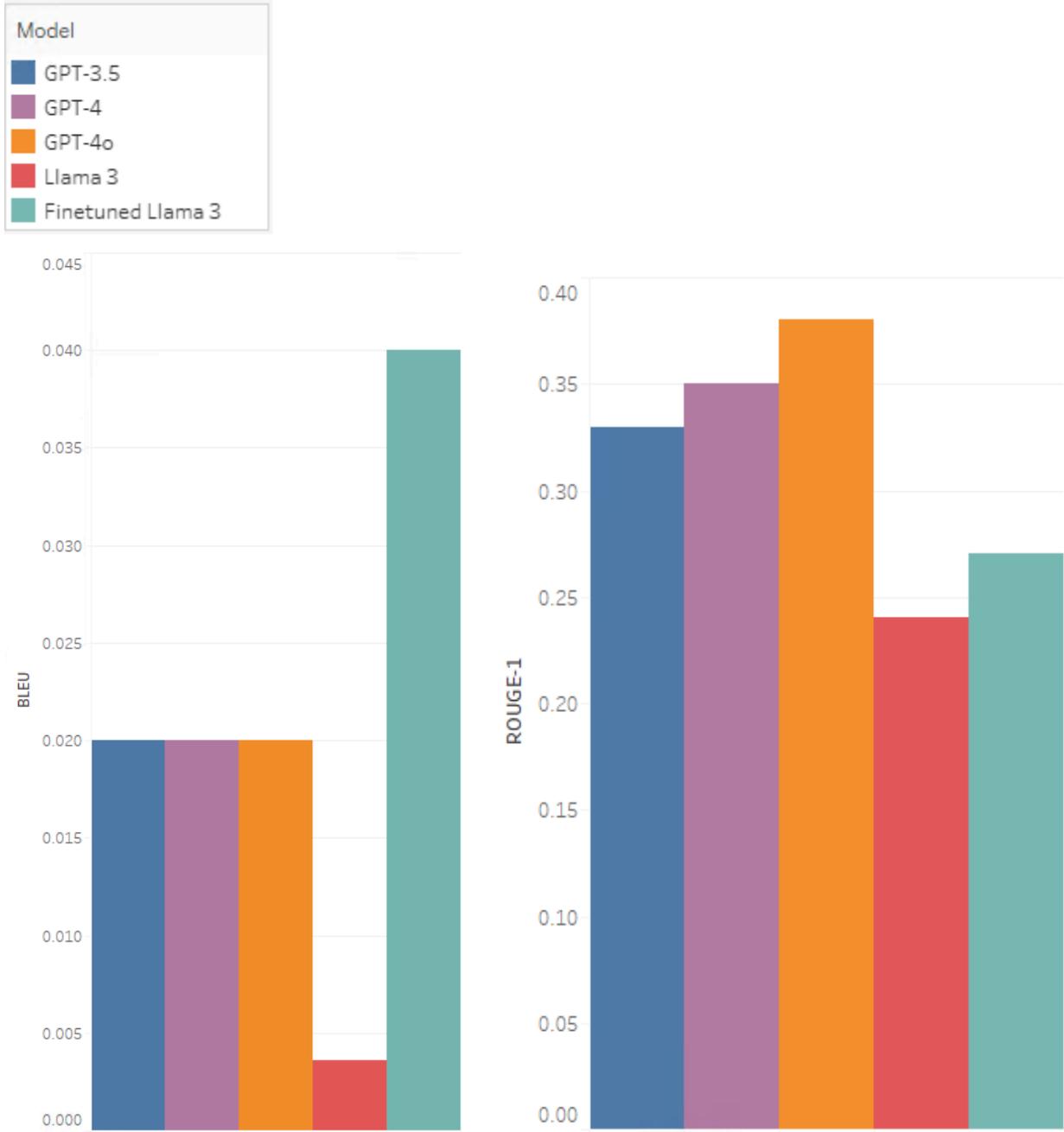

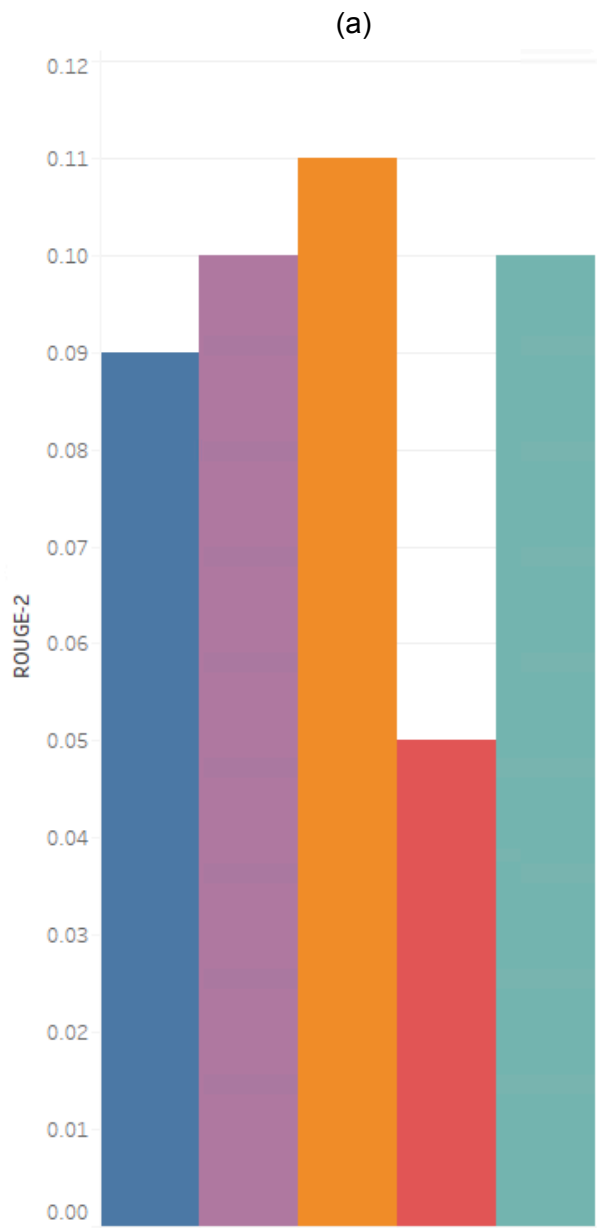
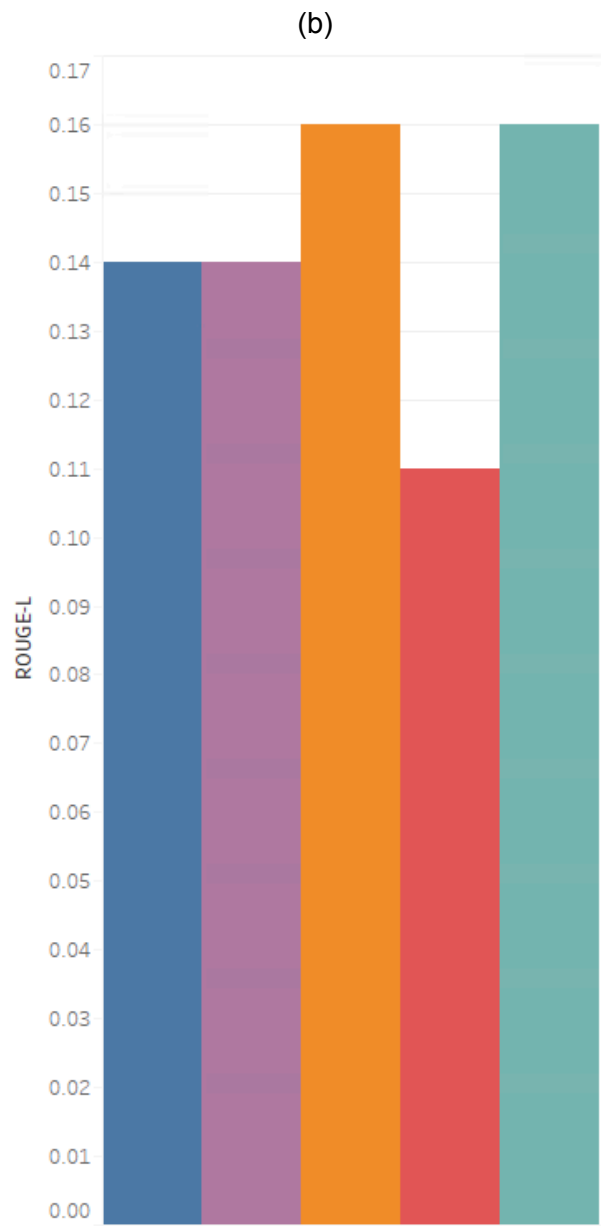

(a) (b)

(c) (d)

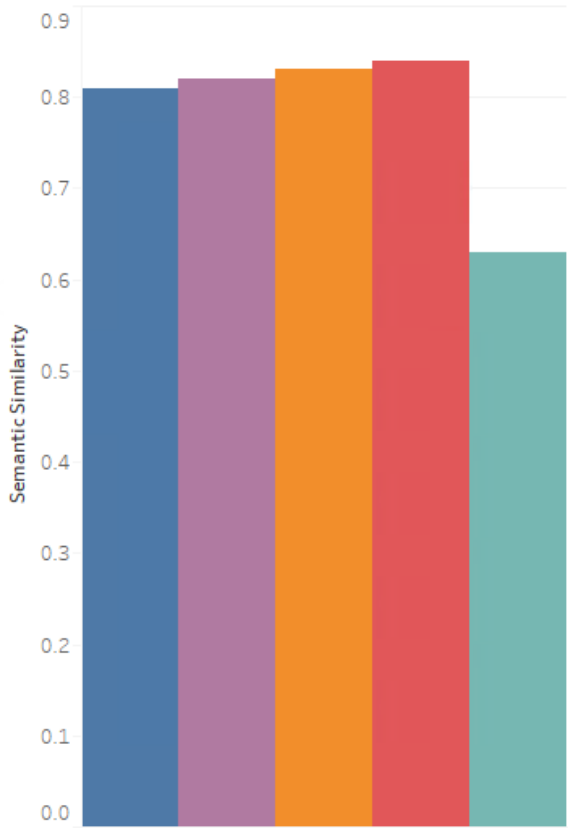

(e)

Figure 3 (a) BLEU (b) ROUGE-1 (c) ROUGE-2 (d) ROUGE-L (e) semantic similarity of GPT-3.5, GPT-4, GPT-4o, LLaMA 3, and fine-tuned LLaMA 3.

Table 1 provides a comparative analysis of the average and standard deviation of token counts for input notes, gold standard summaries, and outputs from LLaMA 3 across different token count ranges.

For input notes, the average number of tokens increases significantly across the ranges, from 13,935.634±11,297.677 for the <50,000 range to 975,395.059±702,735.716 for the >500,000 range, highlighting the variability and extensive length of clinical notes. In contrast, the gold standard summaries remain relatively consistent in length, averaging between 1,552.499±1,190.753 and 2,425.980±1,880.113 tokens, even as the input note length increases.

The LLaMA 3 model's outputs show remarkable consistency in token count across all ranges, with average token counts ranging from 583.286±72.303 to 591.911±138.889. This consistency indicates LLaMA 3's ability to generate concise summaries regardless of the length of the input notes.

Overall, Table 1 highlights the substantial reduction in token count achieved by LLaMA 3 in summarizing lengthy clinical notes, maintaining a concise output even as the input complexity increases.

Table 1 The average and standard deviation of the number of tokens in the input notes, gold standard summaries, and outputs from LLaMA 3 for different token count ranges

| Number of Token Range | Input Notes | Gold Standard | Output |
| --- | --- | --- | --- |
| <50,000 | 13935.634±11297.677 | 1552.499±1190.753 | 591.911±138.889 |
| 50,000-100,000 | 75331.527±14508.694 | 1562.75±836.831 | 583.286±72.303 |
| 100,000-200,000 | 146115.923±29708.030 | 1782.153±1222.929 | 591.302±115.333 |
| 200,000-500,000 | 286892.579±79951.551 | 1823.868±1159.413 | 583.744±64.616 |
| >500,000 | 975395.059±702735.716 | 2425.980±1880.113 | 585.647±67.928 |

Table 2 reports the performance metrics for LLaMA 3 outputs across different token count ranges, including the number of discharge summaries, BLEU, ROUGE-1, ROUGE-2, ROUGE-L, and semantic similarity scores.

The BLEU scores are consistently low across all token ranges, with a slight increase for the >500,000 range (0.004). ROUGE-1 scores remain relatively stable, with the highest score of 0.239 in the 100,000-200,000 range, and the lowest in the >500,000 range (0.228). ROUGE-2 scores are fairly consistent across all ranges, with a slight increase in the 50,000-10,0000 range (0.051). ROUGE-L scores show minor variations, with the highest score (0.109) in the 50,000-100,000 range and the lowest in the >500,000 range (0.102).

Semantic similarity scores are generally high and stable, with slight variations. The highest semantic similarity score (0.837) is observed in the >500,000 range, while the lowest (0.831) is found in both the 50,000-100,000 and 200,000-500,000 ranges.

Overall, the table indicates that LLaMA 3's performance metrics remain relatively consistent across different token count ranges, with minor variations in BLEU, ROUGE, and semantic similarity scores.

Table 2 The number of discharge summaries, BLEU, ROUGE-1, ROUGE-2, ROUGE-L, and semantic similarity scores for LLaMA 3 outputs across different token count ranges

| Number of Token Range | N | BLEU | ROUGE-1 | ROUGE-2 | ROUGE-L | Semantic Similarity |
|---|---|---|---|---|---|---|
| <50,000 | 415 | 0.003 | 0.235 | 0.048 | 0.106 | 0.832 |
| 50,000-100,000 | 112 | 0.003 | 0.237 | 0.051 | 0.109 | 0.831 |
| 100,000-200,000 | 248 | 0.003 | 0.239 | 0.049 | 0.106 | 0.834 |
| 200,000-500,000 | 121 | 0.004 | 0.233 | 0.047 | 0.104 | 0.831 |
| >500,000 | 51 | 0.004 | 0.228 | 0.048 | 0.102 | 0.837 |

# DISCUSSION

This study aimed to evaluate the effectiveness of LLMs in generating discharge summaries for lung cancer patients based on clinical notes. Our findings reveal several important insights into the capabilities and limitations of these models.

## Token-Level Evaluation vs. Semantic Similarity

Although the large language models, including GPT-3.5, GPT-4, and LLaMA 3, did not achieve high scores in token-level evaluation metrics such as BLEU and ROUGE, they demonstrated strong performance in semantic similarity with the gold standard summaries. Semantic similarity scores were notably high across models, indicating that while the generated summaries may not always match in exact wording or structure, they effectively capture the underlying meaning and context of the clinical information. This capability is crucial in clinical settings where understanding the clinical context and meaning is paramount for accurate decision-making.

In the context of medical summarization, the challenges of token-level evaluation metrics are compounded by the intricacies of clinical language and the diversity of patient presentations. Clinical narratives frequently contain specialized terminology, ambiguous references, and contextual dependencies that defy straightforward word-for-word alignment. LLMs excel in capturing the essence of these narratives by leveraging their vast knowledge base and contextual understanding. Semantic similarity metrics, crucially, evaluate the models' ability to distill complex medical concepts into succinct summaries that retain clinical relevance and accuracy. This alignment with clinical utility transcends mere linguistic precision, focusing instead on conveying the clinically significant information in a format that aids decision-making and continuity of care. The nuanced interpretation and synthesis required in medical summarization underscore the importance of semantic coherence over rigid token-level matching, highlighting a shift in evaluation priorities towards metrics that better reflect the pragmatic demands of healthcare applications.

## Effect of Fine-Tuning on Performance

Fine-tuning LLaMA 3, improved token-level evaluation metrics but did not consistently enhance semantic similarity scores. This observation suggests that while fine-tuning enhances the models' ability to mimic the style and vocabulary of clinical summaries, it may not sufficiently refine their understanding and representation of complex medical concepts.

Several factors contribute to this outcome, including the limited scope of training data, which may not fully capture the variability and complexity of clinical scenarios encountered in practice. Training datasets often prioritize generalizability and volume over the nuanced details of individual patient cases or rare medical conditions. As a result, LLMs trained on such data may struggle to generalize effectively to diverse clinical contexts, where patient-specific variables, disease progressions, and treatment outcomes can significantly influence the content and tone of medical summaries. Moreover, the challenge of summarizing extensive clinical notes into concise and clinically relevant summaries persists due to the intricate nature of medical language and the need for precise terminology that reflects both disease pathology and patient management strategies. Adapting to specific patient conditions, which may involve rare diseases or atypical presentations, poses further challenges, as training datasets may lack sufficient examples to adequately fine-tune models for these specialized cases. Therefore, while LLMs show promise in automating summarization tasks, their effectiveness hinges on the richness and representativeness of the data used for training, which currently remains a critical area for improvement in clinical natural language processing.

## Effect of Context Length on LLaMA 3 8b Performance

We observed that LLaMA 3 8b consistently produced concise summaries across varying lengths of input clinical notes. This consistency underscores LLaMA 3's robust performance in summarization tasks, maintaining clarity and relevance regardless of the input note's complexity or length. The model's ability to condense lengthy clinical narratives into succinct summaries highlights its potential utility in real-world clinical settings where brevity and precision are critical.

The consistent performance of LLaMA 3 8b across different token groups suggests that its architecture effectively manages varying input lengths without sacrificing coherence or accuracy. This adaptability is advantageous in handling diverse clinical datasets and supports its applicability in generating standardized and informative discharge summaries. LLaMA 3's robust performance in maintaining clarity and relevance regardless of the input note's complexity or length underscores its potential utility in real-world clinical settings where brevity and precision are critical. This consistency in producing concise summaries across varying lengths of input clinical notes further highlights LLaMA 3's capability to condense lengthy clinical narratives into succinct and clinically relevant summaries, addressing the inherent challenges of summarizing complex medical information comprehensively yet succinctly. Thus, LLaMA 3 8b stands out not only for its technical robustness but also for its practical suitability in enhancing efficiency and quality in clinical documentation processes.

## Strengths and Limitations, and Future Directions

This study's strength lies in its comprehensive comparison of multiple most advanced LLMs in a clinical summarization task. By examining a range of models such as GPT-3.5, GPT-4, GPT-4o, LLaMA 3, and fine-tuned LLaMA 3 versions, we provided a nuanced comparison of their performance across token-level evaluation metrics like BLEU and ROUGE, as well as semantic similarity scores, offering insights into their capabilities and limitations in summarizing medical information.

However, limitations include the reliance on specific datasets and the need for further exploration in larger and more diverse clinical contexts. The datasets used, while representative of clinical notes and summaries, may not fully capture the breadth of variability and complexity seen across different healthcare settings and patient populations. This raises questions about generalizability, as models trained on limited datasets might exhibit biases or limitations when applied to broader, more diverse clinical scenarios. Additionally, the study primarily focused on discharge summaries for lung cancer patients, which may not fully encompass the range of medical conditions and documentation requirements encountered in broader clinical practice. Future research should therefore aim to incorporate larger and more diverse datasets from various clinical specialties to better assess the robustness and generalizability of these models across different healthcare contexts.

Moreover, while the study provided valuable insights into the performance of fine-tuned models like GPT-4o and LLaMA 3 in enhancing token-level evaluation metrics, the observed challenges in achieving high semantic similarity scores highlight ongoing limitations in current natural language understanding capabilities. The discrepancies between token-level metrics and semantic coherence suggest that while models can generate text that matches token-level criteria, deeper semantic understanding and contextual relevance remain areas requiring further development. Addressing these challenges may involve strategies such as refining training methodologies, incorporating domain-specific knowledge bases, or exploring advanced model architectures that prioritize semantic accuracy alongside syntactic fidelity.

## **CONCLUSION**

This study makes significant contributions to clinical informatics by evaluating the efficacy of large language models in generating concise discharge summaries for lung cancer patients. The findings underscore the models' varying strengths in handling clinical data and highlight LLaMA 3 8b's consistent performance across diverse input lengths. These insights are pivotal for advancing automated summarization tools in healthcare, aiming to improve documentation efficiency and clinical decision-making, ultimately improving patient care and healthcare operations.

# ACKNOWLEDGEMENTS

We extend our sincere appreciation to Prof. Xiaoqian Jiang for his support with the GPT4 experiments in this study.

# REFERENCES

[1]  C. van Walraven and A. L. Weinberg, "Quality assessment of a discharge summary system.," *CMAJ*, vol. 152, no. 9, pp. 1437–1442, May 1995.
[2]  A. J. H. Kind and M. A. Smith, "Documentation of Mandated Discharge Summary Components in Transitions from Acute to Subacute Care," in *Advances in Patient Safety: New Directions and Alternative Approaches (Vol. 2: Culture and Redesign)*, K. Henriksen, J. B. Battles, M. A. Keyes, and M. L. Grady, Eds., in Advances in Patient Safety. , Rockville (MD): Agency for Healthcare Research and Quality (US), 2008. Accessed: Jun. 03, 2024. [Online]. Available: http://www.ncbi.nlm.nih.gov/books/NBK43715/
[3]  S. B. Patel and K. Lam, "ChatGPT: the future of discharge summaries?," *The Lancet Digital Health*, vol. 5, no. 3, pp. e107–e108, Mar. 2023, doi: 10.1016/S2589-7500(23)00021-3.
[4]  Y. Li, Q. Wei, X. Chen, J. Li, C. Tao, and H. Xu, "Improving tabular data extraction in scanned laboratory reports using deep learning models," *Journal of Biomedical Informatics*, vol. 159, p. 104735, Nov. 2024, doi: 10.1016/j.jbi.2024.104735.
[5]  Y. Li *et al.*, "RefAI: a GPT-powered retrieval-augmented generative tool for biomedical literature recommendation and summarization," *Journal of the American Medical Informatics Association*, vol. 31, no. 9, pp. 2030–2039, Sep. 2024, doi: 10.1093/jamia/ocae129.
[6]  Y. Li *et al.*, "Relation extraction using large language models: a case study on acupuncture point locations," *Journal of the American Medical Informatics Association*, vol. 31, no. 11, pp. 2622–2631, Nov. 2024, doi: 10.1093/jamia/ocae233.
[7]  E. Y. Zhang, A. D. Cheok, Z. Pan, J. Cai, and Y. Yan, "From Turing to Transformers: A Comprehensive Review and Tutorial on the Evolution and Applications of Generative Transformer Models," *Sci*, vol. 5, no. 4, Art. no. 4, Dec. 2023, doi: 10.3390/sci5040046.
[8]  N. Zhu, N. Zhang, Q. Shao, K. Cheng, and H. Wu, "OpenAI's GPT-4o in surgical oncology: Revolutionary advances in generative artificial intelligence," *European Journal of Cancer*, vol. 206, Jul. 2024, doi: 10.1016/j.ejca.2024.114132.
[9]  "Introducing Meta Llama 3: The most capable openly available LLM to date," Meta AI. Accessed: Jun. 03, 2024. [Online]. Available: https://ai.meta.com/blog/meta-llama-3/
[10] S. Singh, A. Djalilian, and M. J. Ali, "ChatGPT and Ophthalmology: Exploring Its Potential with Discharge Summaries and Operative Notes," *Seminars in Ophthalmology*, vol. 38, no. 5, pp. 503–507, Jul. 2023, doi: 10.1080/08820538.2023.2209166.
[11] R. A. J. Clough, W. A. Sparkes, O. T. Clough, J. T. Sykes, A. T. Steventon, and K. King, "Transforming healthcare documentation: harnessing the potential of AI to generate discharge summaries," *BJGP Open*, vol. 8, no. 1, Apr. 2024, doi: 10.3399/BJGPO.2023.0116.
[12] D. Dubinski *et al.*, "Leveraging artificial intelligence in neurosurgery—unveiling ChatGPT for neurosurgical discharge summaries and operative reports," *Acta Neurochir*, vol. 166, no. 1, p. 38, Jan. 2024, doi: 10.1007/s00701-024-05908-3.
[13] L. Ruinelli *et al.*, "Experiments in Automated Generation of Discharge Summaries in Italian," in *Proceedings of the First Workshop on Patient-Oriented Language Processing*

(CL4Health) @ LREC-COLING 2024, D. Demner-Fushman, S. Ananiadou, P. Thompson, and B. Ondov, Eds., Torino, Italia: ELRA and ICCL, May 2024, pp. 137–144. Accessed: Jun. 04, 2024. [Online]. Available: https://aclanthology.org/2024.cl4health-1.17

[14] "COVID-19 cases | WHO COVID-19 dashboard." Accessed: Jun. 02, 2024. [Online]. Available: https://data.who.int/dashboards/covid19/cases

[15] "Coronavirus Death Toll and Trends - Worldometer." Accessed: Jun. 02, 2024. [Online]. Available: https://www.worldometers.info/coronavirus/coronavirus-death-toll/

[16] O. P. Mehta, P. Bhandari, A. Raut, S. E. O. Kacimi, and N. T. Huy, "Coronavirus Disease (COVID-19): Comprehensive Review of Clinical Presentation," *Front. Public Health*, vol. 8, Jan. 2021, doi: 10.3389/fpubh.2020.582932.

[17] H. Liu, S. Chen, M. Liu, H. Nie, and H. Lu, "Comorbid Chronic Diseases are Strongly Correlated with Disease Severity among COVID-19 Patients: A Systematic Review and Meta-Analysis," *Aging Dis*, vol. 11, no. 3, pp. 668–678, May 2020, doi: 10.14336/AD.2020.0502.

[18] A. Leiter, R. R. Veluswamy, and J. P. Wisnivesky, "The global burden of lung cancer: current status and future trends," *Nat Rev Clin Oncol*, vol. 20, no. 9, pp. 624–639, Sep. 2023, doi: 10.1038/s41571-023-00798-3.

[19] M. B. Schabath and M. L. Cote, "Cancer Progress and Priorities: Lung Cancer," *Cancer Epidemiology, Biomarkers & Prevention*, vol. 28, no. 10, pp. 1563–1579, Oct. 2019, doi: 10.1158/1055-9965.EPI-19-0221.

[20] M. Iachina, E. Jakobsen, A. K. Fallesen, and A. Green, "Transfer between hospitals as a predictor of delay in diagnosis and treatment of patients with Non-Small Cell Lung Cancer – a register based cohort-study," *BMC Health Serv Res*, vol. 17, no. 1, p. 267, Apr. 2017, doi: 10.1186/s12913-017-2230-3.

[21] T. Oksholm, T. Rustoen, and M. Ekstedt, "Transfer Between Hospitals Is a Risk Situation for Patients After Lung Cancer Surgery," *Cancer Nurs*, vol. 41, no. 3, pp. E49–E55, 2018, doi: 10.1097/NCC.0000000000000497.

[22] F. Rodríguez-Cano *et al.*, "Cost-effectiveness of diagnostic tests during follow-up in lung cancer patients: an evidence-based study," *Transl Lung Cancer Res*, vol. 12, no. 2, pp. 247–256, Feb. 2023, doi: 10.21037/tlcr-22-540.

[23] A. Passaro, C. Bestvina, M. Velez Velez, M. C. Garassino, E. Garon, and S. Peters, "Severity of COVID-19 in patients with lung cancer: evidence and challenges," *J Immunother Cancer*, vol. 9, no. 3, p. e002266, Mar. 2021, doi: 10.1136/jitc-2020-002266.

[24] A. Addeo and A. Friedlaender, "Cancer and COVID-19: Unmasking their ties," *Cancer Treatment Reviews*, vol. 88, p. 102041, Aug. 2020, doi: 10.1016/j.ctrv.2020.102041.

[25] M. Maio *et al.*, "Immune Checkpoint Inhibitors for Cancer Therapy in the COVID-19 Era," *Clinical Cancer Research*, vol. 26, no. 16, pp. 4201–4205, Aug. 2020, doi: 10.1158/1078-0432.CCR-20-1657.

[26] R. EL-Andari, N. M. Fialka, U. Jogiat, B. Laing, E. L. R. Bédard, and J. Nagendran, "Resource allocation during the coronavirus disease 2019 pandemic and the impact on patients with lung cancer: a systematic review," *Interdisciplinary CardioVascular and Thoracic Surgery*, vol. 37, no. 6, p. ivad190, Dec. 2023, doi: 10.1093/icvts/ivad190.

[27] J. Luo *et al.*, "COVID-19 in patients with lung cancer," *Annals of Oncology*, vol. 31, no. 10, pp. 1386–1396, Oct. 2020, doi: 10.1016/j.annonc.2020.06.007.

[28] J. Villena-Vargas *et al.*, "Safety of lung cancer surgery during COVID-19 in a pandemic epicenter," *The Journal of Thoracic and Cardiovascular Surgery*, vol. 164, no. 2, pp. 378–385, Aug. 2022, doi: 10.1016/j.jtcvs.2021.11.092.

[29] A. K. Ganti *et al.*, "Risk factors of SARS-CoV-2 infection and complications from COVID-19 in lung cancer patients," *Int J Clin Oncol*, vol. 28, no. 4, pp. 531–542, 2023, doi: 10.1007/s10147-023-02311-3.

[30] Z. Bakouny *et al.*, "COVID-19 and Cancer: Current Challenges and Perspectives," *Cancer*

*Cell*, vol. 38, no. 5, pp. 629–646, Nov. 2020, doi: 10.1016/j.ccell.2020.09.018.

[31] L. Zhang *et al.*, "Clinical characteristics of COVID-19-infected cancer patients: a retrospective case study in three hospitals within Wuhan, China," *Ann Oncol*, vol. 31, no. 7, pp. 894–901, Jul. 2020, doi: 10.1016/j.annonc.2020.03.296.

[32] K. S. Saini *et al.*, "Mortality in patients with cancer and coronavirus disease 2019: A systematic review and pooled analysis of 52 studies," *Eur J Cancer*, vol. 139, pp. 43–50, Nov. 2020, doi: 10.1016/j.ejca.2020.08.011.

[33] J.-B. Leclère *et al.*, "Maintaining Surgical Treatment of Non-Small Cell Lung Cancer During the COVID-19 Pandemic in Paris," *The Annals of Thoracic Surgery*, vol. 111, no. 5, pp. 1682–1688, May 2021, doi: 10.1016/j.athoracsur.2020.08.007.

[34] A. A. Awad, "Content based Document Retrieval using Content Extraction," vol. 4, no. 2.

[35] S. A. Salloum, A. Almarzouqi, B. Gupta, A. Aburayya, M. R. Al Saidat, and R. Alfaisal, "The Coming ChatGPT," in *Artificial Intelligence in Education: The Power and Dangers of ChatGPT in the Classroom*, A. Al-Marzouqi, S. A. Salloum, M. Al-Saidat, A. Aburayya, and B. Gupta, Eds., Cham: Springer Nature Switzerland, 2024, pp. 3–9. doi: 10.1007/978-3-031-52280-2_1.

[36] J. Li *et al.*, "Mapping vaccine names in clinical trials to vaccine ontology using cascaded fine-tuned domain-specific language models," *J Biomed Semant*, vol. 15, no. 1, p. 14, Aug. 2024, doi: 10.1186/s13326-024-00318-x.

[37] Y. Li *et al.*, "Improving Entity Recognition Using Ensembles of Deep Learning and Fine-tuned Large Language Models: A Case Study on Adverse Event Extraction from Multiple Sources," *arXiv.org*, 2024, doi: 10.48550/arXiv.2406.18049.

[38] Y. Hu *et al.*, "Zero-shot Clinical Entity Recognition using ChatGPT," *arXiv.org*, 2023, doi: 10.48550/arXiv.2303.16416.

[39] G. Yenduri *et al.*, "GPT (Generative Pre-Trained Transformer)— A Comprehensive Review on Enabling Technologies, Potential Applications, Emerging Challenges, and Future Directions," *IEEE Access*, vol. 12, pp. 54608–54649, 2024, doi: 10.1109/ACCESS.2024.3389497.

[40] J. Tian *et al.*, "Assessing Large Language Models in Mechanical Engineering Education: A Study on Mechanics-Focused Conceptual Understanding," 2024.

[41] X. Zhang *et al.*, "Automated Root Causing of Cloud Incidents using In-Context Learning with GPT-4," 2024.

[42] S. N. Hart *et al.*, "Organizational preparedness for the use of large language models in pathology informatics," *Journal of Pathology Informatics*, vol. 14, p. 100338, Jan. 2023, doi: 10.1016/j.jpi.2023.100338.

[43] M. U. Hadi *et al.*, "Large Language Models: A Comprehensive Survey of its Applications, Challenges, Limitations, and Future Prospects," Nov. 2023, doi: 10.36227/techrxiv.23589741.v4.

[44] C. Tao *et al.*, "VaxBot-HPV: A GPT-based Chatbot for Answering HPV Vaccine-related Questions," *Research Square*, p. rs.3.rs, Sep. 2024, doi: 10.21203/rs.3.rs-4876692/v1.

[45] Y. Li *et al.*, "Artificial intelligence-powered pharmacovigilance: A review of machine and deep learning in clinical text-based adverse drug event detection for benchmark datasets," *Journal of Biomedical Informatics*, vol. 152, p. 104621, Apr. 2024, doi: 10.1016/j.jbi.2024.104621.

[46] T. Oura *et al.*, "Diagnostic Accuracy of Vision-Language Models on Japanese Diagnostic Radiology, Nuclear Medicine, and Interventional Radiology Specialty Board Examinations," *medRxiv*, p. 2024.05.31.24308072, Jan. 2024, doi: 10.1101/2024.05.31.24308072.

[47] K. I. Roumeliotis, N. D. Tselikas, and D. K. Nasiopoulos, "Llama 2: Early Adopters' Utilization of Meta's New Open-Source Pretrained Model," *Preprints*, vol. 2023, 2023, doi: 10.20944/preprints202307.2142.v2.

[48] Y. Li, J. Li, J. He, and C. Tao, "AE-GPT: Using Large Language Models to extract adverse

events from surveillance reports-A use case with influenza vaccine adverse events," *PLOS ONE*, vol. 19, no. 3, p. e0300919, Mar. 2024, doi: 10.1371/journal.pone.0300919.
[49] W. Huang *et al.*, "How Good Are Low-bit Quantized LLaMA3 Models? An Empirical Study," 2024.
[50] S. G. Picha, D. A. Chanti, and A. Caplier, "Semantic Textual Similarity Assessment in Chest X-ray Reports Using a Domain-Specific Cosine-Based Metric," in *Proceedings of the 17th International Joint Conference on Biomedical Engineering Systems and Technologies - Volume 1: BIOINFORMATICS*, SciTePress, 2024, pp. 487–494. doi: 10.5220/0012429600003657.